\documentclass[11pt,a4paper]{article}
\usepackage[hyperref]{emnlp-ijcnlp-2019}
\usepackage{times}
\usepackage{latexsym}
\pdfoutput=1

\usepackage{url}
\usepackage{color}
\usepackage{caption}
\usepackage{graphicx, subfig}
\usepackage{multirow}
\usepackage{booktabs}
\usepackage{graphicx}

\aclfinalcopy 

\newcommand\dataname{$R^{3}$}

\title{\dataname{}: A Reading Comprehension Benchmark Requiring Reasoning Processes*}

\author{Ran Wang, Kun Tao, Dingjie Song, Zhilong Zhang, Xiao Ma, Xi'ao Su, Xinyu Dai\\
  Nanjing University \\
  {\tt wangr@smail.nju.edu.cn}
  }

\date{}

\begin{document}
\maketitle
\begin{abstract}
Existing question answering systems can only predict answers without explicit reasoning processes, which hinder their explainability and make us overestimate their ability of understanding and reasoning over natural language.
In this work, we propose a novel task of reading comprehension, in which a model is required to provide final answers and reasoning processes.
To this end, we introduce a formalism for reasoning over unstructured text, namely Text Reasoning Meaning Representation (TRMR).
TRMR consists of three phrases, which is expressive enough to characterize the reasoning process to answer reading comprehension questions.
We develop an annotation platform to facilitate TRMR's annotation, and release the \dataname{} dataset, a \textbf{R}eading comprehension benchmark \textbf{R}equiring \textbf{R}easoning processes.
\dataname{} contains over 60K pairs of question-answer pairs and their TRMRs.
Our dataset is available at: \url{http://anonymous}.
\footnote{Work in progress.}
\end{abstract}

\section{Introduction}
The ability to understand and perform reasoning over natural language is an ultimate goal of artificial intelligence.
The machine reading comprehension task provides a quantifiable and objective way to evaluate systems' reasoning ability, where an answer is sought for question given one or more documents.
To this end, many high-quality and large-scale reading comprehension datasets have been released~\cite{DBLP:conf/emnlp/RajpurkarZLL16, DBLP:conf/emnlp/Yang0ZBCSM18, DBLP:journals/tacl/ReddyCM19, DBLP:conf/naacl/DuaWDSS019}, which lay a good data foundation for QA systems in different scenarios.
In turn, various neural models have also emerged recently~\cite{DBLP:conf/iclr/SeoKFH17, DBLP:conf/iclr/YuDLZ00L18, DBLP:conf/iclr/HuangZSC18, DBLP:conf/emnlp/HuPHL19}, which approach or even surpass human-level performance.

Nonetheless, we may overestimate the ability of current QA systems to understand natural language~\cite{DBLP:conf/emnlp/SugawaraISA18}.
Recent analysis suggests that current models can predict gold answers even when original questions are replaced with nonsensical questions~\cite{DBLP:conf/emnlp/FengWGIRB18}, and that higher accuracy does not ensure more robustness or generalization~\cite{DBLP:conf/emnlp/JiaL17,DBLP:conf/emnlp/WallaceWLSG19}.
We argue that one root cause of these problems is that most current QA tasks only require models to predict gold answers.
But metrics, which only depend on answers, are hard to capture our utmost desiderata of these systems, i.e., the ability to understand and reason over natural language~\cite{DBLP:journals/cacm/Lipton18,DBLP:journals/corr/abs-1910-10045}.
Inspired by human beings, giving the reasoning process shows the examinee's abilities more elaborately and comprehensively than just giving the final answer.
We argue it is more accurately evaluate systems’ capabilities by requiring them to explicitly give the reasoning process.

\begin{figure}
\centering
\includegraphics[width=0.5\textwidth]{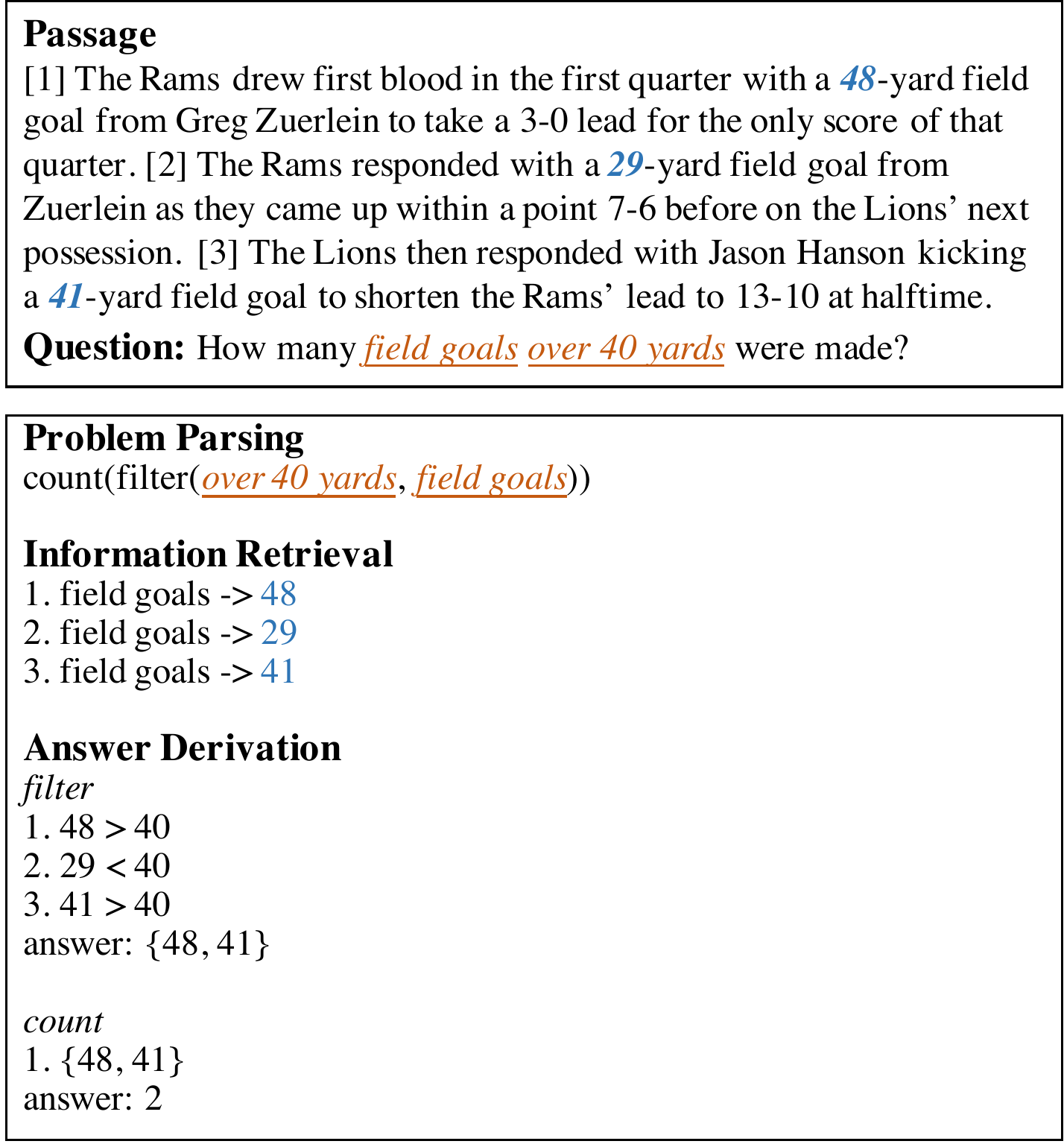}
\caption{
An example from ~\dataname{}.
Each passage-question pair in \dataname{} is annotated with text reasoning meaning representation (TRMR).
The corresponding TRMR is presented, with text spans of passages or questions involved in reasoning colored in orange and in blue for clarity.
}
\label{fig:example}
\end{figure}

In this work, we propose a novel task of reading comprehension over unstructured text, in which a model is required to provide final answers and reasoning processes to demonstrate its ability to understand natural language.
To this end, we construct a large-scale dataset \dataname{}, a \textbf{R}eading comprehension benchmark \textbf{R}equiring \textbf{R}easoning processes.
\dataname{} contains 60k QA pairs, each of which is labeled with an reasoning process.

Annotating reasoning processes precisely across diverse problems is a challenging task even for humans.
To alleviate the difficulty and reduce the cost, we choose to annotate the existing dataset DROP~\cite{DBLP:conf/naacl/DuaWDSS019}, which requires \textit{D}iscrete \textit{R}easoning \textit{O}ver the content of \textit{P}aragraphs.
To do well in DROP, a system must resolve references in a question, perhaps to multiple input positions, and perform discrete reasonings over them (such as addition, counting, or sorting).
These operations force models to understand comprehensively the content of paragraphs.
Furthermore, we propose a formalism for reasoning over unstructured text, namely Text Reasoning Meaning Representation (TRMR), and develop a software to facilitate the annotation task at large scale.

TRMR is inspired by the problem-solving process of human reading comprehension.
When a question needs to be answered, humans first determine the steps required according to the information of passages and questions; then find out the information elements required for the solution; finally, the elements are processed according to the aforementioned determined steps to derive the final answer.
An example of TRMR is shown in Figure~\ref{fig:example}.
Formally, each TRMR contains three steps:
\begin{enumerate}
\item \textit{Problem Parsing} converts questions into atomic operation sequences, where each atomic operation answers sub-questions of the original questions.
For the question ``How many \textit{field goals} \textit{over 40 yards} were made?" in Figure~\ref{fig:example}, it is converted into two predefined atomic operations: ``filter" and ``count".
\item \textit{Information Retrieval} retrieves the items needed to answer those simple questions. For above example, the yards of field goals are retrieved by this step.
\item \textit{Answer Derivation} deduces the final answer according to the reasoning process of ``problem decomposition", the retrieved items of ``information retrieval" and/or answers given by intermediate operations.
\end{enumerate}

Based on the above formalism, we annotate DROP to construct a large-scale reading comprehension dataset, namely \dataname{}.
We make \textit{DN} publicly available at \url{http://anonymous}.

\section{Text Reasoning Meaning Representation}
In this section, we define the \textbf{T}ext \textbf{R}easoning \textbf{M}eaning \textbf{R}epresentation (TRMR).
TRMR aims to characterize the reasoning process when the system answers questions over diverse natural language: determine the problem-solving steps according to passages and questions, find the information needed to answer the questions, and perform operations or reasoning to arrive at the final answer.

\paragraph{TRMR Definition} Formally, given a passage $\mathbf{P} = [w_1, w_2, \cdots, w_{n-1}, w_{n}]$ and a question $\mathbf{Q} = [w_1, w_2, \cdots, w_{l-1}, w_{l}]$,  its TRMR contains three parts: problem parsing, information retrieval and answer derivation.
The ``problem parsing" consists of some predefined operations and the arguments required for these operations, formed as $op_1(op_2(arg_1, arg_2, \cdots), op_3(arg_1, arg_2, \cdots), \cdots)$.
Table~\ref{tab:operations} shows these pre-defined operations.
For simplicity, we restrict these arguments to spans of the question or operations.
Next, ``information retrieval" presents the passage spans needed to answer questions, formed as $arg_1\!\to\!span_1, arg_2\!\to\!span_2, \cdots$, where $span_1, span_2, \cdots$ are spans from passages.
Finally, ``answer derivation" details how to perform operations on the retrieved information based on different operations from ``problem parsing".
An example of TRMR is shown in Figure~\ref{fig:example}.

\begin{table*}[]
\begin{tabular}{ccll}
\toprule
Type &
  Template / Signature &
  \multicolumn{1}{c}{Question} &
  \multicolumn{1}{c}{Problem Parsing} \\ \midrule
\multicolumn{1}{c|}{\multirow{13}{*}{\rotatebox{90}{Arithmetic}}} &
  more($S_1$, $S_2$) &
  \begin{tabular}[c]{@{}l@{}}How many more people were\\ there than households?\end{tabular} &
  more(people, households) \\ \cmidrule{2-4}
\multicolumn{1}{c|}{} &
  more-select($S_1$, $S_2$) &
  \begin{tabular}[c]{@{}l@{}}Who has more people in it,\\Iraq or Iran?\end{tabular} &
  more-select(Iraq, Iran) \\ \cmidrule{2-4} 
\multicolumn{1}{c|}{} &
  less($S_1$, $S_2$) &
  \begin{tabular}[c]{@{}l@{}}How many less households were \\there compared to housing units?\end{tabular} &
  \begin{tabular}[c]{@{}l@{}}less(households,\\ housing units)\end{tabular} \\ \cmidrule{2-4} 
\multicolumn{1}{c|}{} &
  less-select($S_1$, $S_2$) &
  \begin{tabular}[c]{@{}l@{}}Which gender group is smaller:\\ females or male?\end{tabular} &
  less-select(females, male) \\ \cmidrule{2-4} 
\multicolumn{1}{c|}{} &
  cu($S_1$) &
  \begin{tabular}[c]{@{}l@{}}How many percent of people\\ were not white?\end{tabular} &
  cu(white) \\ \cline{2-4} 
\multicolumn{1}{c|}{} &
  completion-more($S_1$) &
  \begin{tabular}[c]{@{}l@{}}How many points were the Bears\\ winning by at halftime?\end{tabular} &
  completion-more(Bears) \\ \cmidrule{2-4}
\multicolumn{1}{c|}{} &
  completion-less($S_1$) &
  \begin{tabular}[c]{@{}l@{}}How many points did the Lions\\ lose the game by?\end{tabular} &
  completion-less(Lions) \\ \cmidrule{2-4}
\multicolumn{1}{c|}{} &
  after($S_1$, $S_2$) &
  \begin{tabular}[c]{@{}l@{}}How many days after the stamps \\arrived were they placed on sale?\end{tabular} &
  \begin{tabular}[c]{@{}l@{}}after(stamps arrived,\\they placed on sale)\end{tabular} \\
  \cmidrule{2-4}
\multicolumn{1}{c|}{} &
  after-select($S_1$, $S_2$) &
  \begin{tabular}[c]{@{}l@{}}What happened second: Poeymirau \\and Freydenberg launched attecks \\or significant riots?\end{tabular} &
  \begin{tabular}[c]{@{}l@{}}after(Poeymirau and Freydenberg \\launched attecks, significant \\riots)\end{tabular} \\
  \cmidrule{2-4}
\multicolumn{1}{c|}{} &
  before($S_1$, $S_2$) &
  \begin{tabular}[c]{@{}l@{}} How many days before the Italians\\ invaded Trieste was the fleet of\\ the Austro-Hungarians destroyed?\end{tabular} &
  \begin{tabular}[c]{@{}l@{}} before(Italians invaded Trieste,\\ fleet of the Austro-\\Hungarians destroyed)\end{tabular} \\
  \cmidrule{2-4}
\multicolumn{1}{c|}{} &
  before-select($S_1$, $S_2$) &
  \begin{tabular}[c]{@{}l@{}} Which happened first, the \\Battle of Vittorio Veneto \\or the Armistice of Villa Giusti?\end{tabular} &
  \begin{tabular}[c]{@{}l@{}} before-select(Battle of \\Vittorio Veneto, Armistice \\of Villa Giusti)\end{tabular} \\
\midrule
\multicolumn{1}{c|}{\multirow{3}{*}{\rotatebox{90}{Aggregate}}} &
  sum($S_1$, $S_2$, ...) &
  \begin{tabular}[c]{@{}l@{}}How many percents of the racial\\ makeup of the county was either\\ Asian or Pacific Islander?\end{tabular} &
  sum(Asian, Pacific Islander) \\ \cmidrule{2-4} 
\multicolumn{1}{c|}{} &
  count($S_1$) &
  \begin{tabular}[c]{@{}l@{}}How many times did Manning\\ throw to Clark?\end{tabular} &
  \begin{tabular}[c]{@{}l@{}}count(times did Manning\\ throw to Clark)\end{tabular} \\ \midrule
\multicolumn{1}{c|}{\multirow{3}{*}{\rotatebox{90}{Select}}} &
  time-span($S_1$) &
  \begin{tabular}[c]{@{}l@{}}How many years did Micheal\\ Tippets The Knot Garden use\\ a classical guitar?\end{tabular} &
  \begin{tabular}[c]{@{}l@{}}time-span(Micheal Tippets\\ The Knot Garden use\\ a classical guitar)\end{tabular} \\ \cmidrule{2-4} 
\multicolumn{1}{c|}{} &
  span($S_1$) &
  \begin{tabular}[c]{@{}l@{}}What event finalized the Lordship\\ of Dernbach being transferred\\ to nassau?\end{tabular} &
  \begin{tabular}[c]{@{}l@{}}span(finalized the Lordship\\ of Dernbach being\\ transferred to nassau)\end{tabular} \\
\midrule
\multicolumn{1}{c|}{\rotatebox{90}{Sort}} &
  sort($S_{superlative}$, $S_1$) &
  \begin{tabular}[c]{@{}l@{}}Which racial group made up the\\ smallest percentage of the population?\end{tabular} &
  sort(smallest, racial group) \\ \midrule
\multicolumn{1}{c|}{\rotatebox{90}{Filter}} &
  filter($S_{condition}$, $S_1$) &
  \begin{tabular}[c]{@{}l@{}}Which groups in percent are\\ larger than 21\%?\end{tabular} &
  filter(larger than 21\%, groups) \\
  \bottomrule
\end{tabular}
\caption{
The predefined operators in TRMR's ``proble parsing".
\label{tab:operations}
}
\end{table*}

\section{Data Collection}
In this section, our annotation pipeline for generating \dataname{} is presented, which consists of three phases.
First, we collection question-answer pairs from existing dataset DROP, a reading comprehension benchmark requiring discrete reasoning over paragraphs.
Second, we crowdsource the TRMR annotation of these question-answer pairs.
Finally, we validate the worker annotations in order to maintain their quality.

\paragraph{Question-Answer Collection}
The passages and questions in \dataname{} are all based on training and validation sets in the existing dataset DROP, while the test set portion of this dataset is hidden.
To encourage annotators to ask complex questions, passages from DROP generally have a narrative sequence of events, and often involve many numbers.
They are usually National Football League (NFL) game summaries and history articles.
As for the quality of questions in DROP, ~\cite{DBLP:conf/naacl/DuaWDSS019} present to works with example questions and workers are only allowed to submit questions that a neural QA model could not solve.
By these settings, questions in DROP are generally difficult, which usually requires complex linguistic understanding and discrete reasoning.
We allow interested readers to read ~\citet{DBLP:conf/naacl/DuaWDSS019}.

\paragraph{TRMR annotation}
Annotation TRMRs precisely across diverse problems can be a challenging and time consuming tasks for humans.
To facilitate annotation and standardize the annotation process, we design and develop an annotation platform.
Our platform has the following properties:
(a) corresponding to TRMR, the system frames annotation processes into three steps and enforces the annotators to perform annotation step by step.
(b) to reduce human input errors and improve annotation efficiency, it automatically calcualte the position of spans in questions or passages and generate (possible) answer derivation steps.
(c) it employs quality control strategies.

\textbf{Annotation Platform}
The annotators are provided with a passage, a question and an answer.
They are required to annotate the corresponding TRMR, i.e. ``problem parsing", ``information retrieval" and ``answer derivation" in turn.

\begin{itemize}
    \item \textbf{Problem Parsing} The annotators are instructed to parse questions into reasoning processes according passages and questions.
To prevent having noisy parsing, they can only choose operations from the pre-defined operation sets or select spans in questions as valid arguments.
    \item \textbf{Information Retrieval} After parsing the questions, the list of arguments of operations, i.e., spans in question are presented to annotators. They need to retrieve information from the passage to arrive at answers. Similarly, they are only allowed to annotate the text spans in the passage, rather than manually entering information to avoid errors.
    \item \textbf{Answer Derivation} To improve the annotation efficiency, the system automatically generate the ``answer derivation" part based on existing problem parsing and retrieved information. Annotators can make modifications on this basis to reduce manual error.
\end{itemize}

\paragraph{Worker Validation}
To ensure worker quality, we initially train and dynamically evaluate annotators through a collection of quality-control strategies.
First, we train our annotators to ensure they understand our annotation principles and how to use the annotation platform.
In addition, they are evaluated through a pre-defined set of test questions.
If their accuracy does not reach a certain threshold, they have to be retrained to continue their annotation.

To further evaluate the quality, we conduct random validation to check whether the TRMR annotation are valid or not.
According to this strategy, at least 2 out of 3 validators should assign the TRMR annotation as valid for it to be selected.
The validation accuracy is 95.92\% across different operations.


\section{Related Work}
\paragraph{Question Answering Dataset}
In recent years, there have been more and more large-scale reading comprehension datasets proposed.
Among them, the most well-known is the Stanford Question Answering Dataset (SQuAD)~\cite{DBLP:conf/emnlp/RajpurkarZLL16}, which is constructed based on Wikipedia and through crowdsourcing annotation.
Recently, more data sets have been proposed to evaluate the performance of QA systems in specific scenarios.
CoQA~\cite{DBLP:journals/tacl/ReddyCM19} and QuAC~\cite{DBLP:conf/emnlp/ChoiHIYYCLZ18} are introduced to evaluate how reading comprehension models aggregate information and answer questions in the context of a conversation.
\citet{DBLP:conf/acl/ZhengHS19} propose a large-scale Chinese reading comprehension dataset, ChID, to study the comprehension a unique language phenomenon in Chinese.
Besides, recent works like HotpotQA~\cite{DBLP:conf/emnlp/Yang0ZBCSM18}, RACE~\cite{DBLP:conf/emnlp/LaiXLYH17}, WikiHop~\cite{DBLP:journals/tacl/WelblSR18}, etc., require the ability of multi-step reasoning.
In addition, some datasets require models that can handle common sense~\cite{DBLP:journals/corr/abs-1810-12885, DBLP:conf/naacl/TalmorHLB19}, understand multiple languages~\cite{DBLP:conf/emnlp/CuiCLQWH19}, or have the ability to apply specialized knowledge~\cite{DBLP:journals/corr/abs-1911-12011}.
The emergence of these data sets lays the data foundation for the design of data-hungry models, such as neural networks, and also provides a public benchmark for evaluating QA systems in different scenarios.

In contrast, our data set is annotated based on DROP~\cite{DBLP:conf/naacl/DuaWDSS019}, which focuses on examining the numerical reasoning capabilities under complex language phenomena.
Besides, unlike most previous work, which uses metrics such as EM or F1 to evaluate the final answer, we require the model to explicitly output the reasoning process in order to force them better understanding the document.

\paragraph{Explainable Question Answering}
Although neural network models have achieved promising results on many reading comprehension tasks~\cite{DBLP:conf/iclr/SeoKFH17, DBLP:conf/iclr/YuDLZ00L18, DBLP:conf/iclr/HuangZSC18, DBLP:conf/emnlp/HuPHL19}, some research work points out that we may overestimate the ability of models to understand or reason.
\citet{DBLP:conf/emnlp/JiaL17} shows that reading comprehension models are susceptible to adversarial samples.
\citet{DBLP:conf/emnlp/KaushikL18} have pointed out that using only passages or questions, reading comprehension models can perform surprisingly well.
\citet{DBLP:conf/acl/MinWSGHZ19} reveals that even highly compositional questions can be answered with a single hop if they target specific entity types, or the facts needed to answer them are redundant.
These results show that it is difficult to construct a reading comprehension dataset that really requires multi-step inference and accurately evaluate the performance of the model.

Since our work is based on the DROP dataset, which requires models to perform symbolic reasoning on numbers, we argue that it is easier to avoid models to solve problems by matching.
However, DROP only judges the performance of the model based on the metrics of F1/EM, which may be not enough to fully describe the model's understanding or reasoning ability~\cite{DBLP:journals/corr/abs-1910-10045}.
Therefore, when evaluating models, we not only ask them to give the final answer, but more importantly, to express the intermediate reasoning process explicitly.

\paragraph{Question Decomposition}
Multi-step reasoning in reading comprehension has been a key challenge in QA.
To solve this challenge, some models decompose a compositional question into simpler sub-questions that can be answered by off-the-shelf single-hop reading comprehension models~\cite{DBLP:conf/naacl/TalmorB18, DBLP:conf/acl/MinZZH19, DBLP:journals/corr/abs-2002-09758}.
This decomposition technique can not only improve model performance, but also provide some explainable evidence for its decision making in the form of sub-questions.
Most recently, \citet{DBLP:journals/corr/abs-2001-11770} introduce a Question Decomposition Meaning Representation (QDMR) for questions and release the BREAK dataset.
Our annotation data (problem analysis part) can be easily converted into a problem decomposition format.
Different to BREAK, we also provide inference processes that reach the final answers.

\section{Conclusion}
In this work, we present ~\dataname{}, a large-scale reading comprehension dataset in which a QA system is required to give answers to questions over diverse natural language, but also needed to present the reasoning processes.
We hope this dataset can facilitating the development of explainable QA systems.



\bibliography{emnlp-ijcnlp-2019}
\bibliographystyle{acl_natbib}

\appendix


\end{document}